\pdfoutput=1

\documentclass[11pt]{article}

\usepackage{EACL2023}
\usepackage{times}
\usepackage{latexsym}
\usepackage{longtable}
\usepackage{supertabular}
\usepackage{array}
\usepackage{multicol}

\usepackage[T1]{fontenc}
\usepackage[utf8]{inputenc}

\usepackage{microtype}

\usepackage{inconsolata}

\usepackage{booktabs}
\usepackage{tabularx}

\usepackage{graphicx}

\title{The Detection and Understanding of Fictional Discourse}

\author{Andrew Piper \\
  McGill University \\
  \texttt{andrew.piper@mcgill.ca} \\\And
  Haiqi Zhou \\
  McGill University \\
  \texttt{haiqi.zhou@mail.mcgill.ca} \\}

\begin{document}
\maketitle
\begin{abstract}
In this paper, we present a variety of classification experiments related to the task of fictional discourse detection. We utilize a diverse array of datasets, including contemporary professionally published fiction, historical fiction from the Hathi Trust, fanfiction, stories from Reddit, folk tales, GPT-generated stories, and anglophone world literature. Additionally, we introduce a new feature set of word ``supersenses'' that facilitate the goal of semantic generalization. The detection of fictional discourse can help enrich our knowledge of large cultural heritage archives and assist with the process of understanding the distinctive qualities of fictional storytelling more broadly. 

\end{abstract}

\section{Introduction}

Written fiction continues to play an important role in the lives of readers. Despite predictions about the end of the book \cite{phillips2019does} or the death of the novel \cite{aravamudan2011refusing}, fiction remains a central medium of communication that contributes to a sense of meaning, joy and imagination on the part of readers over their life-spans, from childhood to old age.

Written fiction can take many forms. It can be found within a multi-billion dollar industry of major publishing houses (``big fiction,'' \citet{sinykin2023big}), self-published books by amateur writers, vibrant online communities devoted to fan fiction, independent presses seeking to promote more diverse styles and voices, or major digital cultural heritage archives.

The computational detection of fictional discourse -- identifying whether a text is telling an imaginary story versus a true one -- can be a useful task for two reasons. First, it can help locate fiction within large, unmarked digitized cultural heritage collections, thus enriching our knowledge about the past \cite{underwood2020noveltm, bagga2022hathi, hamilton2023multihathi}. 

Second, the use of predictive modeling can also facilitate the identification of distinctive qualities of fictional discourse thereby highlighting its potential value to readers and society. While prior theoretical work has made strong claims about the absence of distinctiveness surrounding fictional discourse \cite{searle1975logical, currie1990nature}, computational research in this area has shown that fictional discourse represents a very coherent and historically consistent set of linguistic practices \cite{underwood2014understanding, underwood2019distant, piper2016fictionality}.

In this paper, we seek to further contribute to knowledge about the detectability and understanding of fictional discourse. Prior work has focused on using lexical features (i.e. bag-of-words) \cite{underwood2014understanding, underwood2019distant} and LIWC features \cite{piper2016fictionality} to detect fictional documents within large and small collections of historical texts respectively. In this paper, we expand the number of data sets used to test the accuracy of our models beyond prior work. In addition to using historical collections derived from the Hathi Trust Digital Library \cite{bagga2022hathi}, we also include data from contemporary professional publishing \cite{piper2022conlit}, fan fiction, social media \cite{ouyang2015modeling}, historical folk tales from around the world from Project Gutenberg, along with a collection of contemporary anglophone fiction from non-Western sources. 

Second, where prior work has used either lexical features or LIWC's psycho-linguistic categories to represent model features, here we rely on word ``supersenses'' to capture more general semantic categories of words (see Table 1). Supersenses are derived from Wordnet's taxonomies and generated through the latest model of bookNLP \cite{bamman2021booknlp}. We find that word supersenses serve two important functions for our analysis. 

First, they not only help generalize our understanding of the semantic behavior of fictional discourse beyond individual keywords. Additionally, they help operationalize our understanding of fictional writing as a distinctive form of ``discourse,'' which we define as a socially constructed form of communication \cite{halliday2013halliday, taylor2013discourse, berger2023social}. We assume that fictional discourse is strongly shaped by the distinctiveness of its agents, actions, and worlds, all of which have a strong semantic aspect (though there may be other kinds of distinguishing features that could be the subject of future work). 

Finally, we also explore prediction at the sentence-level using a fine-tuned BERT model to better understand the minimal tokens necessary to distinguish fictional discourse.

In the sections that follow we describe the principal data sets used along with the experimental set-ups employed to better understand both the predictability of fictional discourse and its distinctive qualities as seen from a general semantic viewpoint. We imagine future work could continue to compare these results across more diverse data sets as well as test higher-level formal features. 

\section{Data and Methodology}

\subsection{Data}

In this paper we use the following data sets to run our classification exercises.

\vspace{3mm}

\noindent \textbf{CONLIT}: 2,754 books belonging to 12 different categories split into fiction (1,934) and non-fiction (820) narratives published since 2001 \cite{piper2022conlit}.

\vspace{3mm}

\noindent \textbf{CONLIT\_Page}: The above dataset using 350 sequential tokens sampled from each document.

\vspace{3mm}

\noindent \textbf{CONLIT\_1P\slash3P}: Subsetted by first-person / third-person fiction and ``Memoir'' \slash ``Biography'' for non-fiction.

\vspace{3mm}

\noindent \textbf{HATHI1M}: 1,671,370 randomly sampled pages of English-language prose drawn from the Hathi Trust Digital Library divided between fiction (765,920) and non-fiction (905,450) \cite{bagga2022hathi}.

\vspace{3mm}

\noindent \textbf{HATHI1M\_19C\slash20C}: Pages sampled with an original publication date between 1825-1875 \slash 1875-2000.

\vspace{3mm}

\noindent \textbf{FANFIC}: 9,948 stories sampled from the top 15 most popular fandoms on Archive of Our Own (A03) as of 2020. All texts are between 2,000 and 10,000 words in length.

\vspace{3mm}

\noindent \textbf{REDDIT}: 2,643 stories drawn from sub-reddits that focus on non-fictional storytelling (such as ``What is your scariest real life story?'') \cite{ouyang2015modeling}. 

\vspace{3mm}

\noindent \textbf{FOLK}: 3,136 world folktale collections downloaded from Project Gutenberg.

\vspace{3mm}

\noindent \textbf{WORLDLIT\_EN}: 243 works of Anglophone fiction told in the third-person from three countries: South Africa, Nigeria and India.

\vspace{3mm}

\noindent \textbf{GPT\_Neutral\slash Confounding}: 100 fictional stories generated by GPT-4 using the prompt, "Can you tell me a short story?" (\textit{Neutral}) or prompting for the use of features indicative of non-fictional narratives to tell a fictional story: ``Can you tell me a made-up short story (i.e. not a true story) where you: 1. use a lot of historical dates, 2. talk about social groups rather than individual characters (such as Americans or Germans but use invented groups), and 3. do not describe anyone's personal appearance or their bodies?'' (\textit{Confounding}).

\subsection{Methodology}

All data was processed using the large model of bookNLP \cite{bamman2021booknlp}. We condition on the frequency of supersense types per document normalized by the token count of the document. Table 1 illustrates a sample of supersense types and the top associated tokens.

\begin{table*}[h]
\begin{center}
\begin{tabular}{ll|ll}
\hline
\textbf{Nouns} & \textbf{Top Tokens} & \textbf{Verbs} & \textbf{Top Tokens} \\
\hline
noun.act & way, work, job & verb.body & smiled, wearing, laughed \\
noun.animal & dog, horse, animal & verb.change & began, get, make \\
noun.artifact & door, room, house & verb.cognition & know, think, knew \\
noun.attribute & way, voice, power & verb.communication & said, say, asked \\
noun.body & eyes, head, hand & verb.competition & fight, play, protect \\

\end{tabular}
\end{center}
\caption{Sample of the 40 total supersense types with their most frequent tokens for the fiction subset of CONLIT.}
\label{tab:your_label}
\end{table*}

We then engage in pairwise classification of various combinations of fictional and non-fictional documents using the random forests algorithm, which has been shown to perform well for text-classification purposes \cite{xu2012improved} and exhibits little difference from other textual classifiers \cite{piper2022toward}. For each experiment, we run five-fold cross-validation and report the mean F1. We then extract the top-weighted features for each classifier.

To find out the minimum amount of text required to predict fiction, we train BERT using five datasets randomly sampled from CONLIT consisting of 5,000 passages with lengths of 1-5 sentences. Note that the longer passages are not newly sampled but contain sentences from the shorter sets (i.e. the five-sentence set contains all the sentences in the four-sentence set plus the next sentence). We reference the original BERT paper \cite{devlin2019bert} to select a set of hyperparameters for grid search. 
The result shows that the best-performing hyperparameters are a batch size of 16, a learning rate of 4e-5, and an epoch of 5. 
Each dataset has balanced classes and is divided into a training set of size 3200, a validation set of size 800, and a test set of size 1000 to conduct fine-tuning and evaluation.

\section{Results}

Table 2 provides a snapshot of our document-level classification experiments with the full results reported in the Appendix. In Table 3 we provide an overview of our sentence-level classification results.

\begin{table}[h]
\begin{center}
\begin{tabular}{l|cl}
\hline
\textbf{Dataset} & \textbf{F1} & \textbf{Top Features} \\
\hline
CONLIT & 0.973 &  verb.perception\\
&& noun.body\\
FANFIC & 0.996 & noun.body \\
&&verb.contact \\
HATHI1M & 0.886 & verb.motion \\
&&verb.perception\\
FOLK & 0.991 & noun.plant \\
&&noun.animal\\
WORLDLIT & 0.893 & noun.body \\
&&verb.contact\\
\end{tabular}
\end{center}
\caption{Overview of comparisons with the two most strongly weighted positive predictive features. See Appendix for full list.}
\label{tab:your_label}
\end{table}

\begin{table}[h]
\begin{center}
\begin{tabular}{l|lll}
\hline
\textbf{Sentence Length} & \textbf{Precision} & \textbf{Recall} & \textbf{F1} \\
\hline
One sentence & 0.78 & 0.81 & 0.80 \\
Two sentences & 0.81 & 0.87 & 0.84 \\
Three sentences & 0.85 & 0.89 & 0.87 \\
Four sentences & 0.87 & 0.88 & 0.87 \\
Five sentences & 0.89 & 0.91 & 0.90 \\

\end{tabular}
\end{center}
\caption{BERT Sentence-Level Classification Report}
\label{tab:your_label}
\end{table}

\section{Discussion}

\subsection{Predictibility of Fictional Discourse}
Overall, our work shows that fictional discourse exhibits extremely strong distinguishing features, ranging from a high of an F1 of .99 for fan-fiction and folk-tales to a low of .89 for historical fiction. Even at the sentence level, our BERT-based classifier was able to accurately identify fictional discourse at the single sentence level with at least .80 accuracy that rises to .90 by a five sentence context (though interestingly never achieves the accuracy of our Random Forests model). One of the core insights provided here is the way fictional discourse signals its semantic distinctiveness in very clear and overt ways, such that our classifier has little trouble detecting the difference.

This consistency also exhibits interesting historical continuity at least since the early-nineteenth-century (i.e. the onset of literary Romanticism). As we can see in the Appendix, models trained on nineteenth-century fiction can predict late-twentieth-century fiction as well as the global average for the entire Hathi1M collection. The distinctiveness of fictional writing appears from a semantic point of view to have remained reliably consistent over time.

When we test non-North American fiction, we find that our models lose accuracy but this is due to the smaller training size. If we use a similarly-sized subset of the CONLIT data, we find that the accuracy is comparable. If we test countries individually through a hold-out model (train on WORLDLIT+CONLIT minus one country at a time) we find that the three non-Western collections do exhibit different levels of accuracy, with a low of 0.894 for India and a high of 1.0 for South Africa.  

We also note two interesting results of our GPT-generated stories experiments (see Appendix). When training on contemporary fiction, our models struggle to adequately classify GPT-generated stories, even the simple, straightforward kind (GPT\_Neutral). However, when trained on folk tales, all stories were accurately classified as fiction. This was true even for the ``confounding'' stories which were prompted requesting stories that exhibited high levels of features that were typically indicative of non-fiction.

These results suggest two interesting things: a.) the default GPT theory of ``story'' is highly dependent on the folk-tale genre and b.) stories that exhibit confounding features, i.e. try to sound non-fictional, are still easily detectable as fiction due to other semantic qualities, given the right training data.

\subsection{Distinctive Qualities of Fictional Discourse}
In terms of understanding the distinctive features that our models suggest, we note there is a high-level of consistency with some interesting deviations between datasets.

In CONLIT, the most strongly weighted features for predicting fictional discourse belonged to sensorimotor categories (\textit{verb.perception, noun.body, verb.contact}), which was the same for historical fiction except \textit{verb.motion} replaced \textit{verb.contact}. Fanfiction similarly replicated the same features as CONLIT, but with a lower emphasis on perception. Folktales not surprisingly looked the most anamolous with a far stronger emphasis on noun entities like plants, animals, and food but body language was still important. 

Finally, the WORLDLIT anglophone collection exhibits very similar leading features to CONLIT, suggesting cross-cultural semantic norms in fictional storytelling. 

These findings strongly support prior research that has shown a strong emphasis on embodied behavior on the part of fictional characters as an increasingly core aspect of fictional storytelling over time \citep{piper2023jcls}. Across numerous kinds of datasets and fictional storytelling scenarios, \textit{embodiment} remains a core dimension of constructing and communicating imaginary stories.  

\section{Conclusion}

This paper has shown that much of the earlier findings around fictional discourse's predictability and its emphasis on embodied behavior hold true across a diverse array of corpora. As we discuss in our limitations, future work will want to continue to expand the diversity of genres, languages, and cultures as well move to more non-semantic features to further deepen our understanding of the role that fictional storytelling plays around the world.

\section*{Limitations}
One of the core limitations of our work is the application to English-language texts. While we have expanded the range of text-types, historical periods, and regional cultures used in fictional discourse analysis compared to prior work, future work on multilingual analysis awaits.

Our insights also only pertain to the semantic dimension of fictional discourse. While we find the use of ``supersenses'' a valuable tool for generalizing about semantic behavior, future work will want to focus on the unique formal or structural qualities of fictional discourse when compared to non-fictional narratives.

\section*{Acknowledgmenets}

\bibliography{anthology,custom}
\bibliographystyle{acl_natbib}

\appendix
\section{Appendix}
\label{sec:appendix}

We include figures of the distribution of the top 15 feature weights for three of our primary experiments. Table 4 on next page provides a list of all experiments, F1 scores, and top 5 features.

\begin{figure}
    \resizebox{\linewidth}{7cm}{\includegraphics{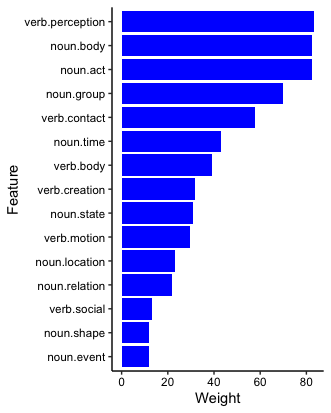}}
    \caption{CONLIT}
    \label{fig:conlit}
\end{figure}
\begin{figure}
    \resizebox{\linewidth}{7cm}{\includegraphics{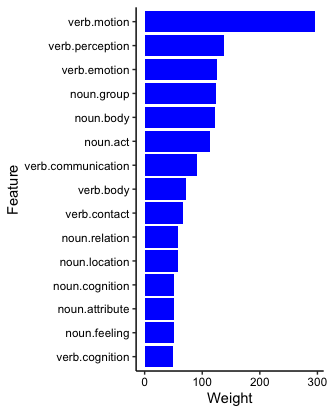}}
    \caption{HATHI1M}
    \label{fig:hathi}
\end{figure}
\begin{figure}
    \resizebox{\linewidth}{7cm}{\includegraphics{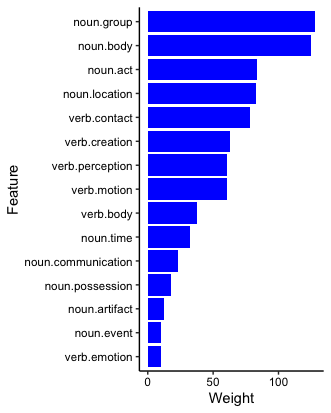}}
    \caption{FANFIC}
    \label{fig:fanfic}
\end{figure}

\newpage
\onecolumn
\tablefirsthead{\hline \textbf{Dataset} & \textbf{F1} & \textbf{Top 5 Features} & \textbf{Support} \\ \hline}
\tablehead{\hline \multicolumn{4}{|c|}{\small\sl continued from previous page} \\ \hline \textbf{Dataset} & \textbf{F1} & \textbf{Top 5 Features} & \textbf{Support} \\ \hline}
\tabletail{\hline \multicolumn{4}{|c|}{\small\sl continued on next page} \\ \hline}
\tablelasttail{\hline}

\topcaption{Full list of experiments undertaken for fiction detection. Note for Train/Test experiments we report ``Accuracy'' not F1 because the test sets are a single class. Bold features are positive predictors of fiction and underline features are positive predictors of non-fiction.}

\begin{supertabular}{p{3.5cm}p{1cm}p{3cm}p{1.5cm}}

&&&\\
CONLIT\_FIC & 0.973 & \textbf{verb.perception} & 1,934 \\
CONLIT\_NON & & \textbf{noun.body} & 627 \\
&& \underline{noun.act} & \\
&& \underline{noun.group} & \\
&& \textbf{verb.contact} & \\
&&&\\
CONLIT\_1P & 0.964 & \underline{noun.act} & 1,025 \\
CONLIT\_MEM & & \underline{noun.time} & 229 \\
&& \textbf{verb.contact} & \\
&& \textbf{noun.body} & \\
&& \underline{noun.group} & \\
&&&\\
CONLIT\_3P & 0.987 & \textbf{verb.perception} & 900 \\
CONLIT\_BIO & & \textbf{noun.body} & 193 \\
&& \textbf{verb.contact} & \\
&& \underline{noun.act} & \\
&& \textbf{verb.body} & \\
&&&\\
FANFIC & 0.996 & \underline{noun.group} & 9,948 \\
CONLIT\_NON & & \textbf{noun.body} &627 \\
&& \underline{noun.act} & \\
&& \underline{noun.location} & \\
&& \textbf{verb.contact} & \\
&&&\\
FANFIC & 0.998 & \textbf{noun.feeling} & 9,948 \\
REDDIT & & \underline{verb.competition} & 2,643 \\
&& \underline{noun.relation} & \\
&& \underline{noun.phenomenon} & \\
&& \underline{noun.possession} & \\
&&&\\
HATH1M FIC & 0.886 & \textbf{verb.motion} & 2,500 \\
HATH1M NON & & \textbf{verb.perception} & 2,500 \\
&& \textbf{verb.emotion} & \\
&& \underline{noun.group} & \\
&& \textbf{noun.body} & \\
&&&\\
HATHI1M\_19C & 0.903 & \textbf{verb.motion} & 2,500 \\
(Train) & & \underline{noun.group} &  \\
HATHI1M\_20C && \textbf{verb.perception} &2,500 \\
(Test) && \textbf{noun.body} & \\
&& \textbf{verb.emotion} & \\
&&&\\
FOLK & 0.991 & \textbf{noun.plant} & 5,000 \\
HATHI1M\_NON & & \textbf{noun.animal} & 3,136 \\
19C&& \textbf{verb.body} & \\
&& \textbf{verb.competition} & \\
&& \textbf{noun.food} & \\
&&&\\
WORLDLIT\_EN & 0.893 & \textbf{noun.body} & 243 \\
CONLIT\_NON & & \underline{noun.act} & 627 \\
&& \textbf{verb.contact} & \\
&& \textbf{verb.perception} & \\
&& \underline{noun.time} & \\
&&&\\
CONLIT\_NYT & 0.901 & \underline{noun.group} & 243 \\
CONLIT\_NON & & \underline{noun.act} & 627 \\
&& \underline{verb.relation} & \\
&& \textbf{verb.perception} & \\
&& \textbf{noun.body} & \\
&&&\\
CONLIT\_PAGE & 0.675 & \textbf{noun.body} & 2,561 \\
(Train)  & & \underline{noun.group} &  \\
GPT\_NEUTRAL && \textbf{verb.contact} & 100 \\
(Test)&& \underline{noun.act} & \\
&& \textbf{verb.motion} & \\
&&&\\
CONLIT\_PAGE & 0.0 & \textbf{noun.body} & 2,561 \\
(Train) & & \underline{noun.group} &  \\
GPT\_CONFOUNDING && \textbf{verb.contact} & 100\\
(Test) && \underline{noun.act} & \\
&& \textbf{verb.motion} & \\
&&&\\
FOLK\slash   & 1.0 & \underline{noun.location} & 3,763 \\
CONLIT\_NON & & \textbf{noun.animal} &  \\
(Train) && \underline{verb.social} & \\
GPT\_NEUTRAL&& \underline{noun.time} & 100\\
(Test)&& \underline{noun.group} & \\
&&&\\
FOLK\& & 0.995 & \underline{noun.location} & 3,763 \\
CONLIT\_NON& & \textbf{noun.animal} &  \\
(Train) && \underline{verb.social} & \\
GPT\_CONFOUNDING&& \underline{noun.time} &100 \\
(Test)&& \underline{noun.group} & \\
&&&\\

WORLDLIT\& & & \textbf{noun.body} & 2,710 \\
CONLIT (Train)& & \underline{noun.act} &  \\
India (Test)& 0.894& \textbf{verb.perception} & 94 \\ 
Nigeria (Test) &0.931& \underline{noun.group} & 72 \\
S. Africa (Test) & 1.0& \textbf{verb.contact} & 77 \\
&&&\\

\end{supertabular}
\end{document}